# Désambiguïsation lexicale à base de connaissances par sélection distributionnelle et traits sémantiques

Mokhtar Boumedyen Billami[1]
(1) Aix-Marseille Université, LIF UMR 7279, 163 avenue de Luminy, 13288 Marseille Cedex 9
mokhtar.billami@lif.univ-mrs.fr

**Résumé.** La désambiguïsation lexicale permet d'améliorer de nombreuses applications en traitement automatique des langues (TAL) comme la recherche d'information, l'extraction d'information, la traduction automatique, ou la simplification lexicale de textes. Schématiquement, il s'agit de choisir quel est le sens le plus approprié pour chaque mot d'un texte. Une des approches classiques consiste à estimer la similarité sémantique qui existe entre les sens de deux mots puis de l'étendre à l'ensemble des mots du texte. La méthode la plus directe donne un score de similarité à toutes les paires de sens de mots puis choisit la chaîne de sens qui retourne le meilleur score (on imagine la complexité exponentielle liée à cette approche exhaustive). Dans cet article, nous proposons d'utiliser une méta-heuristique d'optimisation combinatoire qui consiste à choisir les voisins les plus proches par sélection distributionnelle autour du mot à désambiguïser. Le test et l'évaluation de notre méthode portent sur un corpus écrit en langue française en se servant du réseau sémantique BabelNet. Le taux d'exactitude obtenu est de 78% sur l'ensemble des noms et des verbes choisis pour l'évaluation.

**Abstract.**
**A Knowledge-Based Approach to Word Sense Disambiguation by distributional selection and semantic features.**
Word sense disambiguation improves many Natural Language Processing (NLP) applications such as Information Retrieval, Information Extraction, Machine Translation, or Lexical Simplification. Roughly speaking, the aim is to choose for each word in a text its best sense. One of the most popular method estimates local semantic similarity relatedness between two word senses and then extends it to all words from text. The most direct method computes a rough score for every pair of word senses and chooses the lexical chain that has the best score (we can imagine the exponential complexity that returns this comprehensive approach). In this paper, we propose to use a combinatorial optimization metaheuristic for choosing the nearest neighbors obtained by distributional selection around the word to disambiguate. The test and the evaluation of our method concern a corpus written in French by means of the semantic network BabelNet. The obtained accuracy rate is 78 % on all names and verbs chosen for the evaluation.

**Mots-clés :** désambiguïsation lexicale non supervisée, mesure de similarité distributionnelle, mesures de similarité sémantique.

## 1 Introduction

La désambiguïsation des sens de mots est une *« tâche intermédiaire »* (Wilks, Stevenson, 1996), qui ne constitue pas une fin en soi, mais est plutôt nécessaire à un niveau ou à un autre pour accomplir la plupart des tâches de traitement des langues. Ainsi, la désambiguïsation lexicale suscite de l'intérêt depuis les premiers jours du traitement informatique de la langue (Ide, Véronis, 1998). Elle est nécessaire pour plusieurs applications telles que la recherche d'information, l'extraction d'information, la traduction automatique, l'analyse du contenu, la fouille de textes, la lexicographie et le web sémantique. La plupart des systèmes de désambiguïsation lexicale existants s'appuient sur deux grandes étapes (Navigli, 2009) : (1) représentation de l'ensemble des sens d'un mot ; et (2) choix du sens le plus proche du mot par rapport à son contexte. La première étape repose sur l'utilisation de ressources lexicales telles que les dictionnaires ou les réseaux sémantiques. Ide et Véronis (1998) ont montré que la meilleure possibilité d'identifier le sens d'un mot ambigu est de se référer à son contexte.

L'une des approches les plus classiques pour déterminer le sens le plus probable d'un mot polysémique est d'estimer la proximité sémantique entre chaque sens candidat par rapport à chaque sens de chaque mot appartenant au contexte du mot à désambiguïser[1]. En d'autres termes, il s'agit de donner des scores locaux et de les propager au niveau global. Une application de cette méthode est proposée dans (Pederson *et al.*, 2005). On imagine la complexité exponentielle que retourne cette approche exhaustive. On se retrouve facilement avec un temps de calcul très long alors que le contexte qu'il est possible d'utiliser est petit. Par exemple, pour une phrase de 10 mots avec 10 sens en moyenne, il y aurait $10^{10}$ combinaisons possibles. Le calcul exhaustif est donc très compliqué à réaliser et, surtout, rend impossible l'utilisation d'un contexte plus important. Pour diminuer le temps de calcul on peut utiliser une fenêtre autour du mot afin de réduire le temps d'exécution d'une combinaison mais le choix d'une fenêtre de taille quelconque peut mener à une perte de cohérence globale de la désambiguïsation.

Le contexte du mot à désambiguïser est délimité par une fenêtre textuelle qui se situe à gauche ou à droite ou des deux côtés et dont la taille peut varier. Les fenêtres peuvent être délimitées soit à l'aide de séparateurs de phrases ou de paragraphes, soit à l'aide de *« n-grammes »* qui permettent d'observer un certain nombre (*n-1*) de mots entourant le mot polysémique dans le texte. La définition de la taille de la fenêtre textuelle est liée à celle de la distance optimale entre les mots ambigus et les indices contextuels pouvant servir à leur désambiguïsation (Audibert, 2007). Selon Yarowsky (1993), une grande fenêtre est nécessaire pour lever l'ambiguïté des noms alors que seulement une petite fenêtre suffit pour le cas des verbes ou des adjectifs. Dans un cadre d'analyse distributionnelle de données, plusieurs recherches sont faites sur la construction automatique de thésaurus à partir de cooccurrences de mots provenant d'un corpus de grande taille. Pour chaque mot cible en entrée, une liste ordonnée de voisins les plus proches (*nearest neighbours*) lui est attribuée. Les voisins sont ordonnés en termes de la similarité distributionnelle qu'ils ont avec le mot cible. Lin (1998) a proposé une méthode pour mesurer la similarité distributionnelle entre deux mots (un mot cible et son voisin). Dans cet article, nous nous intéressons à cette approche d'analyse distributionnelle et nous l'utilisons dans la tâche de la désambiguïsation lexicale.

Dans un premier temps, nous présentons un état de l'art sur les méthodes de désambiguïsation lexicale (section 2), notre méthodologie de désambiguïsation à base de traits sémantiques est présentée dans la section 3. Elle décrit le corpus de travail et d'évaluation, le réseau sémantique BabelNet que nous utilisons pour le choix des sens de mots, la mesure de similarité distributionnelle pour le choix des voisins les plus proches ainsi que les algorithmes permettant de retourner le sens le plus probable d'un mot selon le contexte dans lequel il apparait. Nous présentons par la suite les données des expériences menées (section 4) ainsi que les résultats de l'évaluation des différents algorithmes (section 5). Nous terminons par une conclusion et quelques perspectives (section 6).

## 2   État de l'art

Nous pouvons distinguer deux grandes catégories de méthodes de désambiguïsation : (1) dirigées par les données, où l'on trouve les méthodes supervisées et non supervisées. Les méthodes supervisées s'appuient sur un corpus d'apprentissage réunissant des exemples d'instances désambiguïsées de mots. Les méthodes non supervisées exploitent les résultats de méthodes automatiques d'acquisition de sens ; (2) basées sur les connaissances, nécessitant une modélisation étendue aux informations lexico-sémantiques ou encyclopédiques. Ces méthodes peuvent être combinées avec les méthodes non supervisées. La désambiguïsation peut être de deux types : (a) désambiguïsation ciblé, seulement sur un mot particulier dans un texte ; (b) désambiguïsation complète pour tous les mots pleins[2] d'un texte. Il y a deux critères importants pour choisir l'algorithme à utiliser. Le premier critère est une mesure de similarité qui dépend des contraintes de la base de connaissances et du contexte applicatif. Le deuxième critère est le temps d'exécution de l'algorithme. Le lecteur pourra consulter (Ide, Véronis, 1998) pour les travaux antérieurs à 1998 et (Navigli, 2009) pour un état de l'art complet.

Une annotation de mots d'un corpus avec des sens désambiguïsés provenant d'un inventaire de sens (ex. WordNet) est extrêmement coûteuse. À l'heure actuelle très peu de corpus annotés sémantiquement sont disponibles pour l'anglais ; à notre connaissance, rien n'existe pour le français. Le consortium de données linguistiques (*Linguistic Data Consortium*[3]) a distribué un corpus contenant approximativement 200 000 phrases en anglais issues du corpus Brown et Wall Street Journal dont toutes les occurrences de 191 lemmes ont été annotées avec WordNet (Ng, Lee, 1996). Le corpus SemCor (Miller *et al.*, 1993) est le plus grand corpus annoté sémantiquement. Il contient 352 textes annotés avec près de 234 000 sens au total. Cependant, ces corpus contiennent peu de données pour être utilisés avec des méthodes statistiques. Ng (1997) estime que, pour obtenir un système de désambiguïsation à large couverture et de haute précision, nous avons probablement besoin d'un corpus d'environ 3,2 millions de mots de sens étiquetés. L'effort humain pour construire un tel corpus d'apprentissage peut être estimé à 27 années pour une annotation d'un mot par minute par personne (Edmonds, 2000). Il est clair qu'avec une telle ressource à portée de main, les systèmes supervisés seraient beaucoup plus performants mais ça ne reste qu'une hypothèse.

---

[1]   Il peut s'agir d'une phrase, d'un paragraphe ou d'un texte brut.

[2]   Mots pleins : noms, verbes, adjectifs et adverbes.

[3]   *Linguistic Data Consortium (LDC)*. https://www.ldc.upenn.edu

Des efforts ont été fournis pour annoter sémantiquement des corpus en utilisant des méthodes de boostrapping. Hearst (1991) a proposé un algorithme (*CatchWord*) pour une classification des noms qui comprend une phase d'apprentissage au cours de laquelle plusieurs occurrences de chaque nom sont manuellement annotées. Les informations statistiques extraites du contexte de ces occurrences sont ensuite utilisées pour lever l'ambiguïté d'autres occurrences. Si une autre occurrence peut être désambiguïsée avec certitude, le système acquiert automatiquement des informations statistiques de ces nouvelles occurrences désambiguïsées, améliorant ainsi ses connaissances progressivement. Hearst indique qu'une première série d'au moins 10 occurrences est nécessaire pour la procédure, et que 20 ou 30 occurrences sont nécessaires pour une haute précision.

Enfin, les méthodes de désambiguïsation lexicale à base de connaissances se composent d'une part de mesures de similarité sémantique locales qui donnent une valeur de proximité entre deux sens de mots et, d'autre part, d'algorithmes globaux qui utilisent ces mesures pour trouver les sens selon le contexte à l'échelle de la phrase ou du texte. Plusieurs solutions, autres que l'algorithme exhaustif, ont été proposées. Par exemple, des approches à base de corpus pour diminuer le nombre de combinaisons à examiner comme la recherche des chaînes lexicales compatibles (Vasilescu *et al.*, 2004) ou encore des approches issues de l'intelligence artificielle comme le recuit simulé[4] (Cowie *et al.*, 1992) ou les algorithmes génétiques (Gelbukh *et al.*, 2003). Pour plus de détails, le lecteur pourra consulter (Tchechmedjiev, 2012).

# 3 Méthodologie

Désambiguïser tous les mots pleins d'un corpus dont le contexte représente un paragraphe est une tâche qui demande beaucoup de temps si on se base sur un algorithme exhaustif simple. La clé de notre approche de désambiguïsation est l'observation des voisins de chaque mot polysémique dans le texte : au lieu de comparer chaque sens d'un mot à désambiguïser avec tous les sens de tous les mots qui se trouvent dans le texte, nous faisons une comparaison uniquement avec les sens des voisins sélectionnés au moyen d'une similarité distributionnelle. D'une part, ces voisins fournissent souvent des indices sur le sens le plus probable d'un mot dans un texte. D'autre part, cela nous permet de diminuer le temps d'exécution de l'algorithme et de ne pas perdre une cohérence au niveau de la désambiguïsation de tous les mots du texte. Il s'agit de garder l'homogénéité des mots afin de retourner le sens le plus spécifique à chaque mot au lieu de retourner le sens le plus général.

## 3.1 Corpus de données

### 3.1.1 Corpus de travail

Nous avons à disposition un ensemble de trois corpus de différents genres. Le premier corpus est une collection de l'agence française de presse (*French press agency*[5]). Le deuxième corpus est une collection d'articles d'un journal local français (*l'EST Républicain*[6]). Le troisième corpus est une collection d'articles issue de la ressource encyclopédique libre, *Wikipédia*[7]. L'ensemble des données de ces trois corpus est décrit dans le tableau 1. Ces différents corpus ont été analysés automatiquement par la chaîne de traitement Macaon[8] (Nasr *et al.*, 2011). Nous avons obtenu 2 754 686 triplets[9] différents de dépendances syntaxiques correspondant à 31 774 noms uniques et 5 421 verbes uniques. Ces triplets sont stockés et indexés après extraction de 12 785 450 cooccurrences.

---

[4] Méthode d'optimisation stochastique classique fondée sur les principes physiques du refroidissement des métaux qui a été appliquée à la désambiguïsation lexicale.

[5] *French press agency* (*AFP*). http://www.afp.com/fr

[6] *L'EST Républicain*. http://www.estrepublicain.fr/

[7] *Wikipédia*, encyclopédie libre sur le web. https://fr.wikipedia.org

[8] Macaon, chaîne de traitement permettant d'effectuer des tâches standard du TAL. http://macaon.lif.univ-mrs.fr

[9] Un triplet de dépendance syntaxique se compose d'une tête, d'un type de dépendance et d'un modificateur. Par exemple, (*recouvrer, suj, regard*) est un triplet extrait de la phrase *« leurs regards recouvraient les eaux du fleuve »*.

| Corpus | Phrases | Tokens |
|---|---|---|
| AFP | 2 041 146 | 59 914 238 |
| EST REP | 2 998 261 | 53 913 288 |
| WIKI | 1 592 035 | 33 821 460 |
| Total | 6 631 442 | 147 648 986 |

Tableau 1 : Données du corpus de travail

### 3.1.2   Corpus d'évaluation

Nous travaillons sur deux corpus différents, corpus IREST[10] contenant 10 textes et un corpus brut[11] contenant 20 textes pour un total de 30 textes. Nous avons 6 235 occurrences de mots (4 139 occurrences de mots pleins) et une moyenne de 208 occurrences (138 occurrences de mots pleins) par texte (cf. section 4, tableau 2). Les textes sont lemmatisés et annotés en parties du discours par Macaon. Le travail de désambiguïsation que nous menons porte sur des unités monolexicales (les expressions polylexicales n'ont pas été prises en compte).

## 3.2   Ressource lexicale BabelNet

BabelNet[12] (Navigli, Ponzetto, 2012) est un réseau sémantique multilingue permettant de fournir des sens et des entités nommées[13]. BabelNet a été créé en intégrant automatiquement la plus grande encyclopédie multilingue - c'est-à-dire Wikipédia – avec WordNet (Fellbaum, 1998). La construction de cette ressource s'est faite en deux grandes étapes : (1) mapping entre Wikipédia et WordNet ; (2) un système de traduction automatique, basé sur l'application de traduction en ligne de Google, pour recueillir une grande quantité de concepts multilingues et de compléter par les traductions manuellement éditées dans Wikipédia. La construction de BabelNet a permis de couvrir les sens manquants dans WordNet. Le résultat est une ressource multilingue qui fournit des entrées lexicalisées multilingues, reliées entre elles avec une grande quantité de relations sémantiques. De la même façon que WordNet, BabelNet regroupe les mots en différentes langues par groupes de synonymes appelés *Babel synsets*. Pour chaque *Babel synset*, BabelNet fournit des définitions textuelles (appelées gloses) en plusieurs langues, obtenues à partir de WordNet et Wikipédia. A partir de la version 2.0 de BabelNet (octobre, 2013) cette ressource intègre non seulement Wikipédia mais aussi Wiktionary, Wikidata, OmegaWiki et Open Multilingual WordNet, une collection de WordNets disponibles dans différentes langues. A la différence de WordNet qui offre une seule définition par sens, BabelNet permet d'offrir plusieurs définitions pour plusieurs langues.

La version 2.0 de BabelNet couvrait 50 langues, y compris toutes les langues européennes. Actuellement, la version 3.0 couvre 271 langues et contient plus de 13 millions de synsets. Chaque Babel synset contient en moyenne 5,5 synonymes. Le réseau sémantique comprend toutes les relations lexico-sémantiques de WordNet (hyperonymie et hyponymie, méronymie et holonymie, antonymie et synonymie, etc.). Pour la langue française, BabelNet contient actuellement 4 120 733 synsets et 174 591 mots polysémiques. Nous avons choisi d'utiliser BabelNet parce qu'il offre un très grand nombre de synsets et couvre plusieurs mots polysémiques par rapport à d'autres ressources lexicales pour le français comme Wolf[14] (Hanoka, Sagot, 2012) qui offre dans sa version bêta 1.0 un ensemble de 59 091 synsets décrits avec des synonymes et des définitions.

---

[10]   Textes standards pour des tests de vitesse de lecture. http://vision-research.eu

[11]   Textes de lecture pour enfants en école primaire.

[12]   BabelNet, ressource lexicale. http://babelnet.org

[13]   Nous nous intéressons à la désambiguïsation des sens sans tenir compte de la présence des entités nommées. Nous considérons un sens comme étant un concept dans le réseau sémantique.

[14]   Wolf est inspiré de WordNet et présente un réseau sémantique libre pour le français. http://alpage.inria.fr/~sagot/wolf.html

BabelNet dans sa version 2.5.1 a été utilisé pour la réalisation d'un système de désambiguïsation et de détection d'entités nommées, Babelfy[15] (Moro *et al.*, 2014). Babelfy obtient de bonnes performances grâce à la structure de BabelNet qui permet l'intégration des sens lexicographiques et d'entités encyclopédiques en un seul réseau sémantique. Nous utilisons la version 2.5.1 pour l'évaluation de nos expériences afin de comparer nos résultats avec ceux retournés par Babelfy.

### 3.3    Similarité distributionnelle

La similarité distributionnelle est une mesure indiquant le degré de cooccurrence entre un mot cible et son voisin apparaissant dans des contextes similaires. Par exemple, dans un premier texte les voisins de *fleuve* peuvent être *rivière, eau, affluent*. Le sens le plus probable pour *fleuve* est décrit dans BabelNet par trois définitions.

---
**Sens 1**

(1) *Cours d'eau naturel* ;
(2) *En hydrographie, une rivière est un cours d'eau qui s'écoule sous l'effet de la gravité et qui se jette dans une autre rivière ou dans un fleuve, contrairement au fleuve qui se jette, lui, selon cette terminologie, dans la mer ou dans l'océan* ;
(3) *Courant d'eau qui coule d'une altitude élevée à une altitude basse pour arriver dans un lac ou une mer, sauf dans les aires désertiques ou il peut arriver sur rien.*

---

Dans un deuxième texte, les voisins de *fleuve* peuvent être *mer, eau, océan*. Le sens le plus probable pour *fleuve* est décrit dans BabelNet par deux définitions.

---
**Sens 2**

(1) *Cours d'eau se jetant dans une mer* ;
(2) *En hydrographie francophone, un fleuve est un cours d'eau qui se jette dans une mer, dans l'océan, Il se distingue d'une rivière, qui se jette dans un autre cours d'eau.*

---

Plus la similarité distributionnelle entre les voisins est forte plus la probabilité d'avoir le sens le plus probable est grande. Nous utilisons la méthode proposée par Lin (1998) pour l'analyse distributionnelle de données sur notre corpus de travail. Nous avons à disposition un ensemble de relations grammaticales extraites à partir d'une analyse automatique sur les données du corpus de travail. Cette extraction est limitée à un certain nombre de relations de dépendances syntaxiques et nous a permis d'avoir un ensemble de cooccurrences entre les noms et les verbes. Pour chaque nom, nous avons des relations de cooccurrences, par exemple la relation *objet-de* et *sujet-de* ; pour les verbes, la relation *a-pour-objet*, etc. Ainsi, nous avons un ensemble de triplets de cooccurrences $< w, r, x >$ associés avec leur fréquence d'apparition où $r$ est une relation grammaticale et $x$ est une cooccurrence associée avec $w$ selon la relation $r$. Par exemple, les triplets de dépendances syntaxiques dans la phrase *« leurs regards recouvraient les eaux du fleuve »* retournés par la chaîne de traitement Macaon sont : *(regard det leur), (recouvrer suj regard), (recouvrer obj eau), (eau det le), (eau dep de)* et *(de obj fleuve)*[16]. Nous pouvons voir les triplets comme des traits syntaxiques : pour le triplet *(recouvrer, suj, regard)*, *regard* a pour trait syntaxique *suj (recouvrer)*. La similarité distributionnelle entre deux mots $w_1$ et $w_2$ est définie par la fonction suivante :

$$sim(w_1, w_2) = \frac{2x\, I(F(w_1) \cap F(w_2))}{I(F(w_1)) + I(F(w_2))}$$

$F(w_1)$ et $F(w_2)$ représentent l'ensemble des traits syntaxiques possédés respectivement par $w_1$ et $w_2$. $F(w_1) \cap F(w_2)$ représente l'ensemble des traits syntaxiques communs de $w_1$ et $w_2$. Si $I(S)$ est la quantité d'information contenue dans l'ensemble des traits de $S$ alors $I(S) = - \sum_{f \in S} \log P(f)$ où $P(f)$ est la probabilité d'avoir le trait syntaxique $f$. Cette similarité est bornée entre 0 et 1. Elle retourne 1 si $w_1$ et $w_2$ partagent les mêmes traits et retourne 0 si les deux mots n'ont aucun trait en commun. La probabilité $P(f)$ est estimée par le pourcentage des mots qui possèdent le trait syntaxique $f$ parmi l'ensemble des mots possédant la même partie de discours du mot analysé. Sur un ensemble de 30% sélectionné aléatoirement depuis la base de triplets et pour lequel nous avons obtenu 22 168 noms différents, la probabilité d'avoir le trait syntaxique *suj (border)* est de $\frac{38}{22\,168}$ parce que seulement 38 noms uniques sont utilisés comme *sujet* pour le verbe *border*. La quantité d'information pour ce trait est 6.37. Si on prend l'exemple précédant du nom **fleuve**, ce nom possède le trait *suj (border)* comme il possède le trait *obj (connaître)*. La probabilité d'avoir *obj (connaître)* est de $\frac{582}{22\,168}$. La quantité d'information retournée est 3.64. Dans ce cas, le trait *suj (border)* est plus informatif que le trait *obj (connaître)*.

---

[15]   Babelfy, un système de désambiguïsation et de détection d'entités nommées. http://babelfy.org

[16]   La relation *det* est spécifique à un nom et son déterminant ; *suj* est la relation entre un verbe et son sujet ; *obj* est la relation entre un verbe et son objet ou autres ; la dernière relation est *dep* pour présenter une relation générique par défaut.

## 3.4 Similarités sémantiques

Pour mesurer la similarité sémantique, nous utilisons l'algorithme de Lesk (1986) et ses variantes proposées il y a près de 30 ans. Cet algorithme est très simple, il considère la similarité entre deux sens comme le nombre de mots, simplement les suites de caractères séparées par des espaces, en commun dans leurs définitions. La partie 3.4.1 présente l'algorithme de base de Lesk, la partie 3.4.2 présente une variante de Lesk que nous utilisons comme baseline et la partie 3.4.3 présente l'algorithme de Lesk étendu.

### 3.4.1 Algorithme de base de Lesk

Cet algorithme nécessite un dictionnaire (BabelNet pour notre cas) et aucun apprentissage. Il consiste à donner un score à une paire de sens de deux mots différents sans tenir compte ni de l'ordre des mots dans les définitions de ces sens ni d'informations morphologiques ou syntaxiques. Nous faisons une comparaison à partir des traits sémantiques (mots pleins) de chaque définition de sens. Nous utilisons TreeTagger[17] pour obtenir ces traits sémantiques dans notre programme. Comme décrit ci-dessus, BabelNet permet d'offrir plusieurs définitions à un sens pour une langue donnée (français pour nos expériences), nous prenons en compte toutes les définitions possibles. Dans le cas où aucune définition n'est proposée pour un sens, nous prenons en considération les synonymes liés avec le mot à comparer. La fonction utilisée pour mesurer la similarité sémantique se présente par : $Sim_{Lesk}(S_1, S_2) = |D(S_1) \cap D(S_2)|$.

### 3.4.2 Variante de Lesk

Cette variante consiste à retourner le nombre de mots communs entre les unités lexicales (mots pleins) du contexte du mot à désambiguïser et les traits sémantiques des définitions de chaque sens candidat. Navigli (2009) décrit cette variante. Dans nos expériences, le contexte représente le paragraphe. La fonction utilisée pour mesurer la similarité sémantique se présente par : $Lesk_{Variante} = |contexte(w) \cap D(S_i(w))|$ où $w$ est le mot à désambiguïser et $S_i$ est l'IIème sens du mot $w$. Un problème important dans la mesure de Lesk est qu'elle est très sensible aux mots présents dans les définitions. Une absence des mots importants dans les définitions retourne des résultats qui ne sont pas de très bonne qualité. L'une des améliorations proposées pour ce problème est l'algorithme de Lesk étendu.

### 3.4.3 Algorithme de Lesk étendu

Banerjee et Pedersen (2002) proposent la mesure de Lesk étendu. Cette mesure consiste à calculer le recouvrement entre les mots des définitions des deux sens à comparer mais aussi les mots des définitions issues de différentes relations : *hypernyms, hyponyms, meronyms, holonyms et attribute, similar-to, also-see*[18]. Cette mesure est symétrique : une paire de relation $(R_1, R_2)$ est conservée si et seulement si la paire inverse $(R_2, R_1)$ est présente. Un ensemble de relations possibles est obtenu. Si une relation retourne plusieurs sens, toutes les définitions de ces sens sont conservées. Le recouvrement se calcule par la somme des carrés des longueurs de toutes les séquences de mots de la définition *A* dans la définition *B*. La fonction utilisée pour mesurer la similarité se présente par :

$$Sim_{LeskEtendu}(S_1, S_2) = \sum_{\forall (R_1, R_2) \in Relations^2} |D(R_1(S_1)) \cap D(R_2(S_2))|$$

## 3.5 Notre approche

Afin de trouver le sens d'un mot dans un paragraphe, nous utilisons d'abord la mesure de similarité distributionnelle de Lin (1998) pour déterminer un score entre le mot cible (mot à désambiguïser) et l'ensemble des mots du paragraphe qui appartiennent à la même catégorie grammaticale du mot cible. Cela a pour but de retourner les *k* meilleurs voisins qui ont le plus grand score. Ce calcul repose sur les cooccurrences extraites à partir du corpus de travail. Par la suite nous adaptons une méthode structurelle fondée sur une distance sémantique entre les sens selon une formule proposée par Navigli (2009) :

$$S^* = argmax_{S \in \text{Sens}(w)} \sum_{N_i \in N_w: N_i \neq w} max \quad \text{Score}(S, S')$$

---

[17] TreeTagger, outil d'annotation morphosyntaxique. http://www.cis.uni-muenchen.de/~schmid/tools/TreeTagger

[18] Toutes ces relations sont couvertes dans BabelNet.

$S' \in \text{Sens}(N_i)$ avec $i = 1 \ldots k$ et $N_w = \{N_1, N_2, \ldots, N_k\}$ est l'ensemble ordonné des $k$ voisins les plus proches du mot cible $w$. Sens($N_i$) est l'ensemble des sens du voisin $N_i$ et Sens($w$) est l'ensemble des sens du mot cible $w$. Score (S, S') est la fonction utilisée pour mesurer la similarité entre deux sens S et S'. Nous utilisons les deux algorithmes présentés ci-dessus (algorithme de base de Lesk et algorithme de Lesk étendu) et nous comparons notre approche par rapport à la variante de Lesk et Babelfy.

Au niveau de la comparaison des définitions de sens de mots, on peut facilement se retrouver avec des définitions trop concises et il est difficile d'obtenir des distinctions de similarité fines. Pour les trois algorithmes utilisés, nous nous servons de l'heuristique suivante une fois on obtient le score final de chaque sens candidat : « dans le cas où deux sens ou plus possèdent le score de similarité le plus grand, le sens retourné est celui qui a le plus grand nombre de connexions sémantiques avec les autres sens du réseau ». Nous obtenons cette information directement dans BabelNet grâce à sa représentation graphique. Il est mentionné que le sens d'un mot qui a le plus de connexions sémantiques est le plus important. Par exemple, si on prend un extrait d'un texte :

> « ... Leurs regards recouvraient les eaux du **fleuve**. Je ne bougeais plus. Ils m'indiquaient l'étoile du bonheur, quand mon ciel se couvrait de cumulo-nimbus. Je me suis installé derrière eux, confortablement, et j'ai regardé moi aussi couler le **fleuve** du silence ... ».

Le sens de *fleuve* dans ce texte est *« un cours d'eau naturel recevant des affluents et qui se jette dans une rivière ou dans un autre fleuve »*. Il ne s'agit pas d'un cours d'eau qui se jette dans un océan ou dans une mer. Le bon sens par cet exemple possède 2 026 connexions sémantiques et l'autre sens possède 107 connexions sémantiques. L'algorithme de Lesk de base et Lesk étendu retournent le bon sens, en revanche la variante de Lesk (LeskVariante) se trompe.

## 4 Données des expériences menées

Le tableau 2 ci-dessous résume d'une part le nombre de mots reconnus ou non dans BabelNet par catégorie (mot plein) d'unité lexicale pour les types (mots différents) et les tokens (ensemble total de mots). D'autre part, les taux de couverture obtenus. La couverture globale présente le rapport entre les mots reconnus dans BabelNet et l'ensemble des mots du corpus d'évaluation. La couverture des mots polysémiques présente le rapport entre les mots polysémiques reconnus dans BabelNet et l'ensemble des mots couverts par BabelNet (l'ensemble des mots monosémiques et polysémiques).

| POS | Mots polysémiques | | Mots monosémiques | | Mots non reconnus | | Nombre total | | % couverture globale | | % couverture mots polysémiques | |
|---|---|---|---|---|---|---|---|---|---|---|---|---|
| | Tokens | Types | Tokens | Types | Tokens | Types | Tokens | Types | **Tokens** | **Types** | **Tokens** | **Types** |
| **Noms** | 1 660 | 590 | 130 | 39 | 51 | 28 | 1 841 | 657 | **97,23** | **95,74** | **92,74** | **93,8** |
| **Verbes** | 1 135 | 327 | 31 | 30 | 99 | 47 | 1 265 | 404 | **92,17** | **88,37** | **97,34** | **91,6** |
| **Adjectifs** | 353 | 164 | 28 | 19 | 165 | 48 | 546 | 231 | **69,78** | **79,22** | **92,65** | **89,62** |
| **Adverbes** | 375 | 68 | 79 | 6 | 33 | 14 | 487 | 88 | **93,22** | **84,09** | **82,6** | **91,89** |
| **Total** | 3 523 | 1 149 | 268 | 94 | 348 | 137 | 4 139 | 1 380 | **91,59** | **90,07** | **92,93** | **92,44** |

Tableau 2 : Taux de couverture du corpus d'évaluation par la ressource lexicale BabelNet

Nous avons la meilleure couverture globale pour les noms sur les tokens et les types. Nous atteignons 97,23% en tokens contre 92,17% pour les verbes et 95,74% en types contre 88,37% pour les verbes. La couverture en tokens des verbes polysémiques est forte[19] par rapport à la couverture des noms polysémiques (97,34% contre 92,74%). Parmi les cas de non reconnaissance restants, les erreurs d'étiquetage morphosyntaxique représentent la quasi-totalité des cas (seulement 69,78% en tokens reconnus pour les adjectifs). Les mots pleins qui ne sont pas reconnus sont très peu fréquents.

## 4.1 Jeu de test

Nous avons choisi les données de notre jeu de test selon leur niveau d'ambiguïté. Nous avons à disposition un corpus d'évaluation pour lequel il est difficile de faire le choix des mots polysémiques selon leur fréquence d'apparition (peu fréquent, fréquent et très fréquent). De ce fait, notre choix porte sur le niveau d'ambiguïté (peu ambigu, ambigu et très ambigu). Nous prenons deux mots polysémiques pour chaque niveau d'ambigüité et cela pour les noms et les verbes. Le tableau 3 ci-dessous résume les informations quantitatives utilisées pour la sélection des mots polysémiques du jeu de test. Nous considérons les mots qui ont moins de 4 sens comme peu ambigus (cf. tableau 3), les mots qui ont entre 4 et 6 sens comme ambigus et les mots qui ont plus de 6 sens comme très ambigus.

| POS | Candidat | Fréquence | Nombre de synsets | Niveau d'ambigüité |
|---|---|---|---|---|
| **Noms** | fleuve | 3 | 3 | peu ambigu |
| | fée | 8 | 3 | |
| | pêcheur | 4 | 4 | ambigu |
| | plante | 15 | 5 | |
| | castor | 4 | 9 | très ambigu |
| | souris | 10 | 9 | |
| **Verbes** | planter | 2 | 3 | peu ambigu |
| | naître | 7 | 3 | |
| | obliger | 2 | 5 | ambigu |
| | taire | 9 | 5 | |
| | troubler | 2 | 7 | très ambigu |
| | parler | 6 | 10 | |

Tableau 3 : Mots polysémiques du jeu de test avec leur fréquence d'apparition et niveau d'ambigüité

# 5 Évaluation

Pour mesurer les performances des différentes méthodes de désambiguïsation, nous utilisons le taux d'exactitude (*accuracy*). L'évaluation de nos méthodes est effectuée sur des données dont la couverture des sens par BabelNet est de 100%. Ce taux d'exactitude est calculé pour chaque mot du jeu de test et pour chaque méthode de désambiguïsation testée. Il présente le rapport entre le nombre d'occurrences correctement désambiguïsées et le nombre total d'occurrences d'un mot. L'ensemble des taux d'exactitude obtenus est résumé dans le tableau 4. Notre jeu de test contient 44 occurrences pour 6 noms et 28 occurrences pour 6 verbes (un total de 72 occurrences sur 12 mots différents). Nous avons affecté manuellement à chaque occurrence le bon sens proposé dans BabelNet. Notre évaluation porte d'une part sur le niveau d'ambiguïté des mots polysémiques, d'autre part, sur la mesure distributionnelle utilisée pour choisir les *k-plus proches*

---

[19] Nous avons une occurrence par verbe monosémique (30/31) et une couverture de 327 verbes polysémiques contre 30 verbes monosémiques.

*voisins* (*k*-PPV). Notre choix s'est porté sur trois valeurs différentes, $k \in \{3, 5, 7\}$. Nous avons choisi aussi de prendre en compte différentes versions du corpus de travail afin de mesurer le degré de confiance de notre approche en sélectionnant aléatoirement une partie de l'ensemble des triplets de dépendances syntaxiques (30%$V_1$ pour une première version, 30%$V_2$, 50%$V_1$ et 50%$V_2$) ou la totalité des triplets de dépendances. Le tableau 4 présente les résultats obtenus en tenant compte d'une première sélection de 30% sur l'ensemble des triplets.

| Jeu de test | LeskBase | LeskÉtendu | LeskVariante | Babelfy |
|---|---|---|---|---|
| fleuve | 100 | 100 | 0 | 100 |
| fée | 0 | 100 | 100 | 87,5 |
| pêcheur | 0 | 0 | 0 | 0 |
| plante | 86,67 | 100 | 80 | 100 |
| castor | 100 | 100 | 100 | 100 |
| souris | 30 | 100 | 0 | 30 |
| planter | 100 | 100 | 100 | 100 |
| naître | 0 | 85,71 | 85,71 | 100 |
| obliger | 50 | 100 | 50 | 100 |
| taire | erreur POS | erreur POS | erreur POS | - |
| troubler | 0 | 0 | 0 | 0 |
| parler | 16,67 | 100 | 16,67 | 50 |

Tableau 4 : Taux d'exactitude obtenus par méthode de Lesk de base, Lesk étendu et par sélection aléatoire de 30% ($V_1$) sur l'ensemble des triplets pour les 5 plus proches voisins et comparaison avec la méthode de Lesk variante et Babelfy sur les données du jeu de test

Les résultats retournés par l'algorithme de Lesk étendu sont intéressants en comparaison avec les autres algorithmes ou ce que Babelfy retourne sur l'ensemble des noms étudiés. Lesk étendu retourne le bon sens pour les noms peu ambigus, Babelfy retourne un sens différent sur une occurrence de *fée* où il a détecté une expression polylexicale *« fée carabosse »*. Pour les noms ambigus, Lesk étendu retourne le bon sens sur toutes les occurrences de *plante* par contre il se trompe sur toutes les occurrences de *pêcheur*. Cela en raison qu'il existe deux sens pour lesquels le score retourné par nos méthodes est le même : (sens 1) *« la pêche est l'activité consistant à capturer des animaux aquatiques dans leur milieu naturel »* ; (sens 2) *« personne dont la profession est d'attraper des poissons »*. Sur un extrait de texte : *« ... il fut recueilli par un vieux **pêcheur** de saumons ... »*, le bon sens de *pêcheur* est le deuxième mais le premier est retourné par nos méthodes vu qu'il possède plus de connexions sémantiques (1 576 contre 355). Pour les noms très ambigus, Lesk étendu ne se trompe pas contrairement à Babelfy. Sur quelques textes décrivant *la souris* comme *« genre d'animaux »*, Babelfy retourne une entité nommée *« MouseHunt »* décrivant un long métrage de Gore Verbinski.

Pour les verbes, il est difficile de juger la sensibilité du taux d'exactitude au niveau d'ambiguïté. D'une part, nous avons des erreurs d'étiquetage (exp. *taire* ne se trouve sur aucun des textes utilisés), d'autre part, le manque des définitions en français dans BabelNet, ce qui permet de retourner dans la plupart des cas le sens le plus fort du verbe étudié dans le réseau malgré l'utilisation des synonymes. Nous remarquons que l'algorithme de Lesk étendu est beaucoup plus régulier par rapport à l'algorithme de base de Lesk ou à la baseline (Lesk variante). Le meilleur taux d'exactitude que nous obtenons sur l'ensemble des mots étudiés est celui retourné par l'algorithme de Lesk étendu (90,91% pour les noms et 57,14% pour les verbes). Lesk étendu est meilleur par rapport à Babelfy et Lesk variante pour la désambiguïsation des noms (taux d'exactitude de 72,73% par Babelfy et 54,55% par Lesk variante) ainsi que pour la désambiguïsation des verbes (taux d'exactitude de 50% par Babelfy et 35,71% par Lesk variante). Dans le tableau 5, nous présentons les résultats obtenus par variation du nombre des voisins les plus proches et par sélection aléatoire ou non (100%) d'un ensemble de dépendances syntaxiques.

| Algorithme de Lesk étendu | Noms | | | | | Verbes | | | | |
|---|---|---|---|---|---|---|---|---|---|---|
| | k=3 | k=5 | k=7 | Moyenne | Écart typeP | k=3 | k=5 | k=7 | Moyenne | Écart typeP |
| Depends30%$V_1$ | **90,91** | **90,91** | 84,09 | 88,64 | 3,21 | 50 | 57,14 | **60,71** | 55,95 | 4,45 |
| Depends30%$V_2$ | 84,09 | 84,09 | 81,82 | 83,33 | 1,07 | 42,86 | 39,29 | **60,71** | 47,62 | 9,37 |
| Depends50%$V_1$ | 75 | **90,91** | 75 | 80,3 | 7,5 | 25 | 28,57 | 28,57 | 27,38 | 1,68 |
| Depends50%$V_2$ | 75 | 75 | 84,09 | 78,03 | 4,29 | 35,71 | 32,14 | **60,71** | 42,85 | 12,71 |
| Depends100% | 84,09 | 77,27 | 77,27 | 79,54 | 3,21 | 32,14 | **60,71** | **60,71** | 51,19 | 13,47 |
| Moyenne | 81,82 | 83,64 | 80,45 | 81,97 | 1,3 | 37,14 | 43,57 | 54,28 | 45 | 7,07 |
| Écart typeP | **6,1** | **6,65** | **3,69** | **3,76** | - | **8,63** | **13,05** | **12,86** | **9,80** | - |

Tableau 5 : Taux d'exactitude obtenus par algorithme de Lesk étendu par ensemble de triplets pour $k$-PPV

Nous remarquons que la variation de l'ensemble des triplets de dépendances syntaxiques apporte des résultats différents pour la désambiguïsation lexicale (différence légère pour les noms mais forte pour les verbes suite au manque des définitions en français malgré l'utilisation des synonymes). Les voisins d'un mot étudié changent à chaque fois où on utilise un ensemble de triplets différent. Pour l'exemple de *plante*, nous avons sur un texte les voisins (*bande, feuille, oiseau*) par sélection de 30%$V_1$ sur l'ensemble des triplets alors que nous obtenons un autre ensemble de voisins (*animal, insecte, oiseau*) par sélection de 30%$V_2$. Pour les noms et sur la variation des $k$ voisins les plus proches, nous obtenons le meilleur taux d'exactitude (90,91%) pour $k \in \{3, 5\}$ par rapport au cas où $k=7$. Cela signifie qu'un petit nombre de voisins est nécessaire pour retourner le bon sens pour les noms, ce qui est tout le contraire pour les verbes où le meilleur taux d'exactitude retourné est atteint lorsque $k = 7$. Les résultats montrent qu'on obtient un bon degré de confiance pour les noms (écart type de 3,76) par contre un degré de confiance faible pour les verbes (écart type de 9,80). Les figures 1 et 2 présentent les résultats obtenus par utilisation de l'algorithme de Lesk étendu respectivement sur les noms et les verbes. Les figures 3 et 4 présentent les résultats obtenus pour les différents algorithmes utilisés et Babelfy sur un ensemble précis de dépendances syntaxiques.

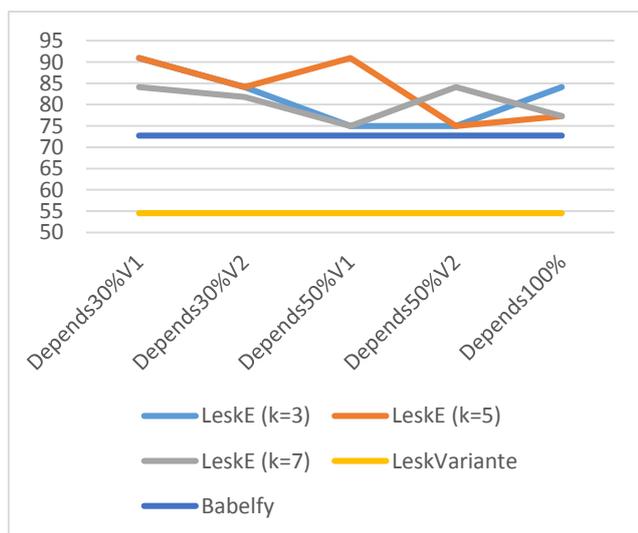 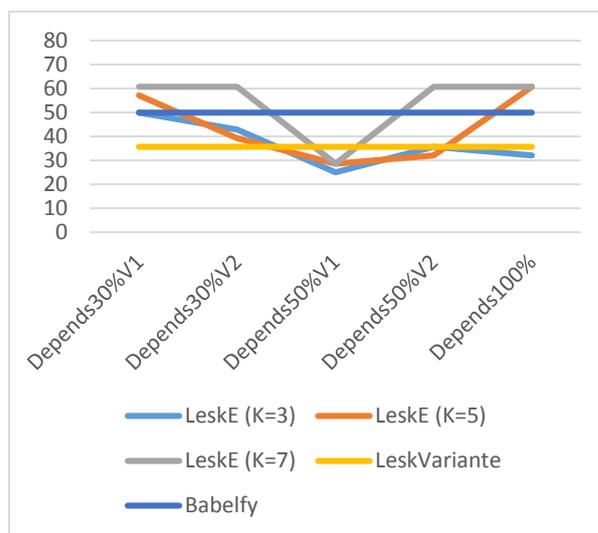

Figure 1: Taux d'exactitude sur les **noms** du jeu de test par utilisation de l'algorithme de Lesk étendu

Figure 2: Taux d'exactitude sur les **verbes** du jeu de test par utilisation de l'algorithme de Lesk étendu

L'algorithme Lesk étendu retourne le meilleur résultat par rapport à la baseline et Babelfy pour les noms et cela sur toutes les variations utilisées pour obtenir un ensemble des triplets de dépendances syntaxiques. Pour les verbes, quelques sélections aléatoires des triplets de dépendances apportent à notre approche des résultats faibles par rapport à la baseline et Babelfy. L'utilisation d'une autre mesure de similarité qui ne repose pas sur des traits sémantiques peut corriger ce problème.

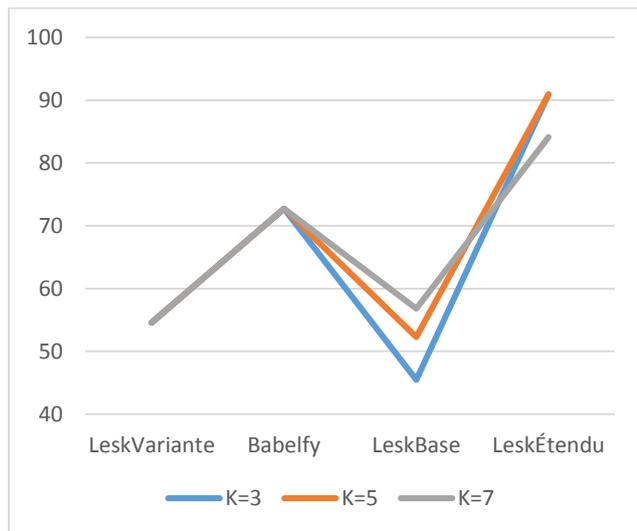 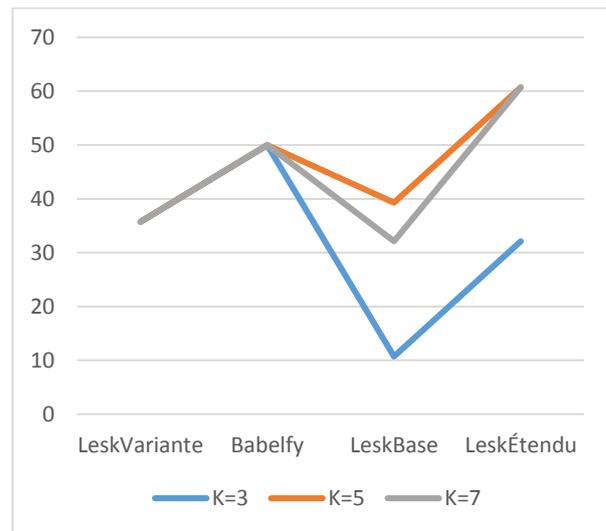

Figure 3: Taux d'exactitude sur les **noms** du jeu de test par sélection aléatoire de 30% sur les dépendances syntaxiques

Figure 4: Taux d'exactitude sur les **verbes** du jeu de test sur l'ensemble des dépendances syntaxiques

## 6   Conclusion et perspectives

Cet article se situe dans le champ de la désambiguïsation lexicale. La méthode que nous avons testée et évaluée s'appuie sur une sélection distributionnelle des voisins les plus proches selon le contexte du mot à désambiguïser. Nous avons adapté l'application d'une approche exhaustive pour se comparer avec *k*-PPV. Cette approche adaptée repose sur l'utilisation de mesures de similarité à base de traits sémantiques. Le contexte utilisé dans cette expérience correspond à un paragraphe et le corpus d'évaluation appartient à un domaine général et non pas à un domaine de spécialité. Le meilleur taux d'exactitude retourné pour les noms est de 90,91% contre 60,71% pour les verbes. La meilleure combinaison retourne 77,78% (k=5 et 30%$V_1$ de l'ensemble des triplets de dépendances syntaxiques).

Sur le plan des perspectives de ce travail, nous envisageons d'utiliser d'autres algorithmes locaux pour mesurer la similarité sémantique. Par exemple, des algorithmes à base de distance taxonomique qui consistent à compter le nombre d'arcs qui séparent deux sens dans une taxonomie (Wu, Palmer, 1994 ; Hirst, St-Onge, 1998) ou des algorithmes à base de contenu informationnel (Resnik, 1995 ; Seco *et al.*, 2004). Nous envisageons aussi de comparer nos résultats avec d'autres algorithmes globaux comme la recherche des chaînes lexicales (Vasilescu *et al.*, 2004) ou les algorithmes génétiques (Gelbukh *et al.*, 2003).

## Remerciements

## Références

Audibert, L. (2007). Désambiguïsation lexicale automatique : sélection automatique d'indices. *Traitement Automatique des Langues Naturelles* (*TALN*)*, Juin 2007, Toulouse, France. IRIT Press,* 13-22.

Banerjee, S., Pedersen, T. (2002). An adapted lesk algorithm for word sense disambiguation using wordnet. *In CICLing '02, London, UK,* 136–145.

Cowie, J., Guthrie, J., Guthrie, L. (1992). Lexical disambiguation using simulated annealing. *In COLING 92, Stroudsburg, PA, USA. ACL,* 359–365.

Edmonds, P. (2000). Designing a task for SENSEVAL-2. Tech. *note. University of Brighton, Brighton. U.K.* Fellbaum, C. Ed. (1998). WordNet: An Electronic Database. *MIT Press, Cambridge, MA.*

Gelbukh, A., Sidorov, G., Han, S. Y. (2003). Evolutionary approach to natural language word sense disambiguation through global coherence optimization. *WSEAS Transactions on Communications,* 1(2):11–19.

Hanoka, V., Sagot, B. (2012). Wordnet creation and extension made simple: A multilingual lexicon-based approach using wiki resources. *In Proceedings of LREC 2012, Istanbul, Turquie.*

Hearst, M. A. (1991). Noun homograph disambiguation using local context in large corpora. *Proceedings of the 7th Annual Conf. of the University of Waterloo Centre for the New OED and Text Research, Oxford, United Kingdom,* 1-19.
Hirst, G., St-Onge, D. D. (1998). Lexical chains as representations of context for the detection and correction of malapropisms. *WordNet : An electronic Lexical Database. C. Fellbaum. Ed. MIT Press,* 305–332.

Ide, N., Véronis, J. (1998). Word sense disambiguation: The state of the art. *Computat. Ling. 24, 1,* 1–40.
Lesk, M. (1986). Automatic sense disambiguation using machine readable dictionaries : how to tell a pine cone from an ice cream cone. *In Proceedings of the 5th annual international conference on Systems documentation, SIGDOC '86, New York, NY, USA : ACM,* 24–26.

Lin, D. (1998). An information-theoretic definition of similarity. *In Proceedings of the 15th International Conference on Machine Learning (ICML, Madison, WI),* 296–304.

Miller, G. A., Leacock, C., Tengi, R., Bunker, R. T. (1993). A semantic concordance. *In Proceedings of the ARPA Workshop on Human Language Technology.* 303–308.

Moro, A., Raganato, A., Navigli, R. (2014). Entity Linking meets Word Sense Disambiguation: a Unified Approach. *Transactions of the Association for Computational Linguistics (TACL), 2,* 231-244.

Nasr, A., Béchet, F., Rey, J. F., Favre, B., Le Roux, J. (2011). MACAON: A linguistic tool suite for processing word lattices. *The 49th Annual Meeting of the Association for Computational Linguistics. ACTI,* 86-91.

Navigli, R., Ponzetto, S. P. (2012). BabelNet: The Automatic Construction, Evaluation and Application of a Wide-Coverage Multilingual Semantic Network. *Artificial Intelligence, 193, Elsevier,* 217-250.

Navigli, R. (2009). Word Sense Disambiguation : a Survey. *ACM Computing Surveys 41(2), ACM Press,* 1-69.
Ng, T. H. (1997). Getting serious about word sense disambiguation. *In Proceedings of the ACL SIGLEX Workshop on Tagging Text with Lexical Semantics: Why, What, and How ? (Washington D.C.),* 1–7.
Ng, H. T., Lee, H. B. (1996). Integrating multiple knowledge sources to disambiguate word sense: An examplar-based approach. *Proceedings of the 34th Annual Meeting of the Association for Computational Linguistics, University of California, Santa Cruz, California,* 40-47.

Pedersen, T., Banerjee, S., Patwardhan, S. (2005). Maximizing Semantic Relatedness to Perform Word Sense Disambiguation. *Research Report UMSI 2005/25, University of Minnesota Supercomputing Institute.*

Resnik, P. (1995). Using information content to evaluate semantic similarity in a taxonomy. *In IJCAI'95, San Francisco, CA, USA,* 448–453.

Seco, N., Veale, T., Hayes, J. (2004). An intrinsic information content metric for semantic similarity in wordnet. *In Proceedings of ECAI'2004, Valencia, Spain,* 1089–1090.

Tchechmedjiev, A. (2012). État de l'art : mesures de similarité sémantique locales et algorithmes globaux pour la désambiguïsation lexicale à base de connaissances. *In Proceedings of the Joint Conference JEP-TALN-RECITAL 2012, volume 3: RECITAL,ATALA/AFCP. June 2012, Grenoble, France,* 295–308.

Vasilescu, F., Langlais, P., Lapalme, G. (2004). Evaluating variants of the lesk approach for disambiguating words. *In Proceedings of LREC 2004, the 4th International Conference On Language Resources And Evaluation, Lisbon, Portugal,* 633–636.

Wilks, Y., Stevenson, M. (1996). The grammar of sense: Is word sense tagging much more than part-of-speech tagging ? *Technical Report CS-96-05, University of Sheffield, Sheffield, United Kingdom.*

Wu, Z., Palmer, M. (1994). Verbs semantics and lexical selection. *In Proceedings of the 32nd annual meeting on ACL, volume 2 de ACL '94, Stroudsburg, PA, USA. Association for Computational Linguistics,* 133–138.

Yarowsky, D (1993). One sense per collocation. *In Proceedings of the ARPA Workshop on Human Language Technology (Princeton, NJ),* 266–271.